\documentclass[journal]{IEEEtran}
\makeatletter
\def\@IEEEfigurecaptionsepspace{-100pt}
\makeatother

\usepackage[linesnumbered,ruled,vlined]{algorithm2e}%[ruled,vlined]{
\usepackage{etoolbox}
\UseRawInputEncoding
\makeatletter
\patchcmd{\@algocf@start}% <cmd>
  {-1.5em}% <search>
  {0pt}% <replace>
  {}{}% <success><failure>
\makeatother

\usepackage{algpseudocode}
\usepackage{setspace}
\SetKwRepeat{Repeat}{repeat}{until}%
\usepackage{bm}
\usepackage{multirow}
\usepackage{url}
\usepackage{tikz}
\usepackage[normalem]{ulem}
\usepackage{array}
\usepackage{cite}
\usepackage{booktabs}
\usepackage{enumitem}
\usepackage{amsfonts}
\usepackage{amssymb}
\usepackage{mathrsfs}
\usepackage{xfrac}
\usepackage{subfigure}
\usepackage{threeparttable}
\usepackage{ragged2e}
\usepackage{graphicx}
\usepackage{threeparttable}
\usepackage{makecell}
\usepackage{diagbox}
\usepackage{color}
\usepackage{bm}
\usepackage{comment}
\usepackage{lipsum}
\usepackage{amsmath}
\usepackage{tabularx}
\usepackage{caption}
\usepackage{arydshln}
\usepackage{tablefootnote}
\usepackage{threeparttable}
\makeatletter
\newcommand*\bigcdot{\mathpalette\bigcdot@{.75}}
\newcommand*\bigcdot@[2]{\mathbin{\vcenter{\hbox{\scalebox{#2}{$\m@th#1\bullet$}}}}}
\algnewcommand{\NULL}{\textsc{null}}
\makeatother
\usepackage{amsthm}
\usepackage{wasysym}
\setlist[itemize,1]{label=$\bigcdot$}
\setlist[itemize,2]{label=-}

\theoremstyle{definition}

\usepackage{setspace}
\ifCLASSINFOpdf
\else
\fi
\begin{document}
\bstctlcite{IEEEexample:BSTcontrol}
\title{
  \fontsize{22pt}{30pt}\selectfont
  \parbox{\linewidth}{\centering
  \textsc{DeepFusion:} Accelerating MoE Training via Federated Knowledge Distillation
  from Heterogeneous Edge Devices
  }
}


\author{Songyuan~Li$^{\,*}$, %~\IEEEmembership{Member,~IEEE,}
        Jia~Hu, %~\IEEEmembership{Member,~IEEE,}
        Ahmed~M.~Abdelmoniem, %~\IEEEmembership{Senior Member,~IEEE}
        Geyong~Min, %~\IEEEmembership{Member,~IEEE,}
        Haojun~Huang, and
        Jiwei~Huang%~\IEEEmembership{Senior Member,~IEEE}
        % Dapeng~Oliver~Wu,~\IEEEmembership{Fellow,~IEEE}% <-this % stops a space
%\thanks{Manuscript received February 10, 2026. (\textit{Corresponding author: Jia Hu; Geyong Min.})}
% This research was supported in part by UKRI-EPSRC Grant No. EP/X035085/1.
\IEEEcompsocitemizethanks{
\IEEEcompsocthanksitem Songyuan~Li and Ahmed~M.~Abdelmoniem are with the School of Electronic Engineering and Computer Science, Queen Mary University of London, London E1 4NS, U.K. (e-mail: S.Y.Li@qmul.ac.uk; ahmed.sayed@qmul.ac.uk).\protect 
\IEEEcompsocthanksitem Jia Hu, and Geyong Min are with the Department of Computer Science, Faculty of Environment, Science and Economy, University of Exeter, Exeter EX4 4PY, U.K. (e-mail: J.Hu@exeter.ac.uk; G.Min@exeter.ac.uk).\protect
\IEEEcompsocthanksitem Haojun Huang is with the School of Electronic Information and Communications, Huazhong University of Science and Technology, Wuhan 430074, China (e-mail: hjhuang@hust.edu.cn).\protect
\IEEEcompsocthanksitem Jiwei Huang is with the Beijing Key Laboratory of Petroleum Data Mining, China University of Petroleum, Beijing 102249, China (e-mail: huangjw@cup.edu.cn).\protect
}
\thanks{* This work was done primarily while Songyuan Li was with the Department of Computer Science, University of Exeter, U.K.}
}

% \markboth{IEEE TRANSACTIONS ON COMPUTERS,~Vol.~XX, No.~X, XXXX~202X}%
% {Shell \MakeLowercase{\textit{et al.}}: Bare Demo of IEEEtran.cls for IEEE Journals}
% \markboth{IEEE TRANSACTIONS ON MOBILE COMPUTING,~Vol.~XX, No.~X, XXXX~202X}%
% {Shell \MakeLowercase{\textit{et al.}}: Bare Demo of IEEEtran.cls for IEEE Journals}



\maketitle

\begin{abstract}
Recent Mixture-of-Experts (MoE)-based large language models (LLMs) such as Qwen-MoE and DeepSeek-MoE are transforming generative AI in natural language processing. However, these models require vast and diverse training data. Federated learning (FL) addresses this challenge by leveraging private data from heterogeneous edge devices for privacy-preserving MoE training. Nonetheless, traditional FL approaches require devices to host local MoE models, which is impractical for resource-constrained devices due to large model sizes. To address this, we propose \textsc{DeepFusion}, the first scalable federated MoE training framework that enables the fusion of heterogeneous on-device LLM knowledge via federated knowledge distillation, yielding a knowledge-abundant global MoE model.  Specifically, \textsc{DeepFusion} features each device to independently configure and train an on-device LLM tailored to its own needs and hardware limitations. Furthermore, we propose a novel View-Aligned Attention (VAA) module that integrates multi-stage feature representations from the global MoE model to construct a predictive perspective aligned with on-device LLMs, thereby enabling effective cross-architecture knowledge distillation. By explicitly aligning predictive perspectives, VAA resolves the view-mismatch problem in traditional federated knowledge distillation, which arises from heterogeneity in model architectures and prediction behaviors between on-device LLMs and the global MoE model. Experiments with industry-level MoE models (Qwen-MoE and DeepSeek-MoE) and real-world datasets (medical and finance) demonstrate that \textsc{DeepFusion} achieves performance close to centralized MoE training. Compared with key federated MoE baselines, \textsc{DeepFusion} reduces communication costs by up to 71$\%$ and improves token perplexity by up to 5.28$\%$.
\end{abstract}


% \fi
\IEEEpeerreviewmaketitle
\section{Introduction}
We have been witnessing rapid advancements in large language models (LLMs)~\cite{Touvron2023LLaMA, GPT4_Tech_Report, DeepSeek_R1_Report}, whose transformer-based architectures and vast text training have revolutionized generative AI for understanding and generating human language in natural language processing (NLP). Building on this momentum, LLMs are driving further advances in NLP performance by embracing novel Mixture-of-Experts (MoE) architectures~\cite{Liu2024DiversifyingMoE}. The MoE-based LLM employs a unique architecture of sparsely activated expert subnetworks, each specializing in distinct domain knowledge to handle diverse input tasks efficiently. Through a gating mechanism, each input is dynamically routed to the relevant expert subnetwork(s), allowing for customized processing and optimal performance. This MoE architecture massively scales model parameters to tens or hundreds of billions (e.g., Qwen3-MoE 235B \cite{Qwen_MoE} and DeepSeek-V2 236B \cite{DeepSeek_MoE}). As a result, the MoE-based LLMs require much larger and more diverse training data to ensure that all expert subnetworks are exposed to diverse and extensive training samples \cite{Jan2024Scaling}, thereby fully leveraging the scaling benefits of MoE. This surging demand exacerbates concerns about the imminent shortage of high-quality public text data for LLM training, as recent projections warn that public human text data for training could be depleted between 2026 and 2032~\cite{Pablo2024Will}.

Federated learning (FL) addresses this challenge by sourcing extensive and diverse training data from ubiquitous edge devices for MoE training, while preserving data privacy. Local model training occurs at distributed edge devices (e.g., smartphones and embedded devices) to obtain an optimal global model through aggregation, without centralizing private on-device data. Traditional FL methods, represented by FedAvg \cite{McMahan2017FedAvg}, restrict on-device training to a local copy of the global model, ensuring the same model architecture and size across devices to enable element-wise averaging at the central aggregation server. Nonetheless, standard FedAvg-like approaches are impractical to implement federated MoE training, as resource-constrained edge devices usually cannot accommodate a full local copy of the global MoE model. To tackle this issue, recent studies~\cite{Guo2021PFL, Dun2023Fedjets, Zhan2024FedMoE, Feng2025PMMOE} have made some attempts in federated MoE training. Specifically, these works introduce a small local expert model deployed at each edge device. The local expert model, which is essentially a compact MoE network pruned from the global MoE, shares its backbone MoE architecture and remains structurally compatible with the global model. Consequently, local expert models can be easily merged back into the global MoE model by aligning their components with those of the global model and assembling them to form the complete global MoE architecture.

However, these studies~\cite{Guo2021PFL, Dun2023Fedjets, Zhan2024FedMoE, Feng2025PMMOE} rely on an impractical assumption that all participating edge devices have sufficient hardware capacity to operate the assigned local expert model. In practice, although these local expert models are much smaller than the global MoE, they remain too cumbersome for edge devices with limited resources, as they still retain the complex MoE backbone and expert gating mechanisms. For instance, smartphones (e.g., iPhone 16) and embedded devices (e.g., NVIDIA Jetson Nano) are popular Artificial Intelligence of Things (AIoT) hardware for on-device LLMs \cite{Zheng2025Review}. However, embedded devices with low-power CPUs/GPUs and limited SRAM/storage have lower AI capabilities compared to smartphones with advanced GPUs/NPUs \cite{Kwon2024TinyTrain}. Consequently, the local expert model could run smoothly on smartphones but often struggles on resource-constrained embedded devices, limiting the applicability of current federated MoE training methods. Given the widespread yet diversified edge devices, it is crucial to enable broad device participation in federated MoE training. This exposes the large-scale MoE model to diverse and extensive training samples, enriching each expert subnetwork's domain knowledge and ultimately enhancing overall MoE performance across varied input tasks.

Therefore, we propose \textsc{DeepFusion}, a novel federated MoE training framework based on knowledge distillation. It enables ubiquitous resource-constrained edge devices with heterogeneous training data to participate in federated MoE training. Distinguished from prior research, \textsc{DeepFusion} allows each edge device to independently configure its on-device LLMs based on local application requirements and resource constraints. These on-device LLMs are trained on local application data to achieve optimal performance, and serve as repositories of local AI knowledge for transfer into the global MoE model. To address FL communication overhead, \textsc{DeepFusion} adopts a communication-efficient FL setting. Once local training is sufficient, the trained LLMs are uploaded in a single round to the central FL server for knowledge transfer. The central server clusters the collected on-device LLMs into local knowledge domains, and integrates them into the global MoE model through federated knowledge distillation. However, we identify a \textit{view-mismatch} problem in this process. The on-device LLMs (teachers) and the global MoE model (student) differ in model architectures, latent spaces, and predictive perspectives (i.e., different inductive biases and preferences when making predictions). This renders traditional knowledge distillation methods based on direct teacher-student logit alignment or feature alignment within a shared latent space ineffective. To address this, we propose a View-Aligned Attention (VAA) module that enables the student model to achieve a similar perspective with that of the teacher model, thereby facilitating efficient knowledge transfer. Finally, the resulting MoE model consolidates expertise from heterogeneous devices, yielding a robust, knowledge-abundant global model. This paper makes the following contributions:
\begin{itemize}[leftmargin=*]
\item To the best of our knowledge, we are the first to source extensive and diverse training data from ubiquitous resource-constrained edge devices for MoE training while preserving data privacy. We propose a novel, scalable federated MoE training framework named \textsc{DeepFusion}. With federated knowledge distillation, on-device LLMs trained on local data serve as repositories of local AI knowledge that can be transferred into the global MoE model.
\item We address the view-mismatch problem in traditional federated knowledge distillation from on-device LLMs to the global MoE model, arising from heterogeneity in model architectures and predictive perspectives. By integrating multi-stage feature representations from the global MoE model (student), the proposed VAA module aligns the student's predictive perspective with that of on-device LLMs (teachers), enabling effective cross-architecture knowledge transfer.
\item \textsc{DeepFusion} is evaluated through comprehensive experiments on two industry-level MoE-based LLM models Qwen-MoE and DeepSeek-MoE, using diverse real-world datasets with data privacy concerns including medical and financial data, across multiple-choice and open-ended question-answering (QA) tasks. Results show that \textsc{DeepFusion} closely matches centralized MoE training and outperforms federated MoE baselines, with token perplexity improvements of up to 5.28$\%$ on challenging open-ended QA tasks and communication cost reductions of up to 71$\%$ across different settings.
\end{itemize}

The remainder of this paper is organized as follows. Section II introduces the preliminaries on MoE-based LLM architecture and their operating mechanism. Section III presents the \textsc{DeepFusion} system design and key implementation challenges. Section IV details the proposed methodology to address these system challenges. Section V evaluates the performance of \textsc{DeepFusion} through comprehensive experiments and ablation studies. Section VI reviews related work, and Section VII concludes the paper.

\begin{comment}
The rest of this article is organized as follows. Section \ref{sec:related_work} introduces the related work. Section \ref{sec:system_model} presents the system model. Section \ref{sec:problem_statement} formulates the multi-dimensional optimization problem. Section \ref{sec:auction_mechanism_design} describes our proposed \texttt{AERIA} mechanism in details. Section \ref{sec:evaluation} discusses the experimental results. Section \ref{sec:conclusion} concludes this article.
\end{comment}

\begin{figure}[t]
\centering
\includegraphics[width= 3.5in, keepaspectratio]{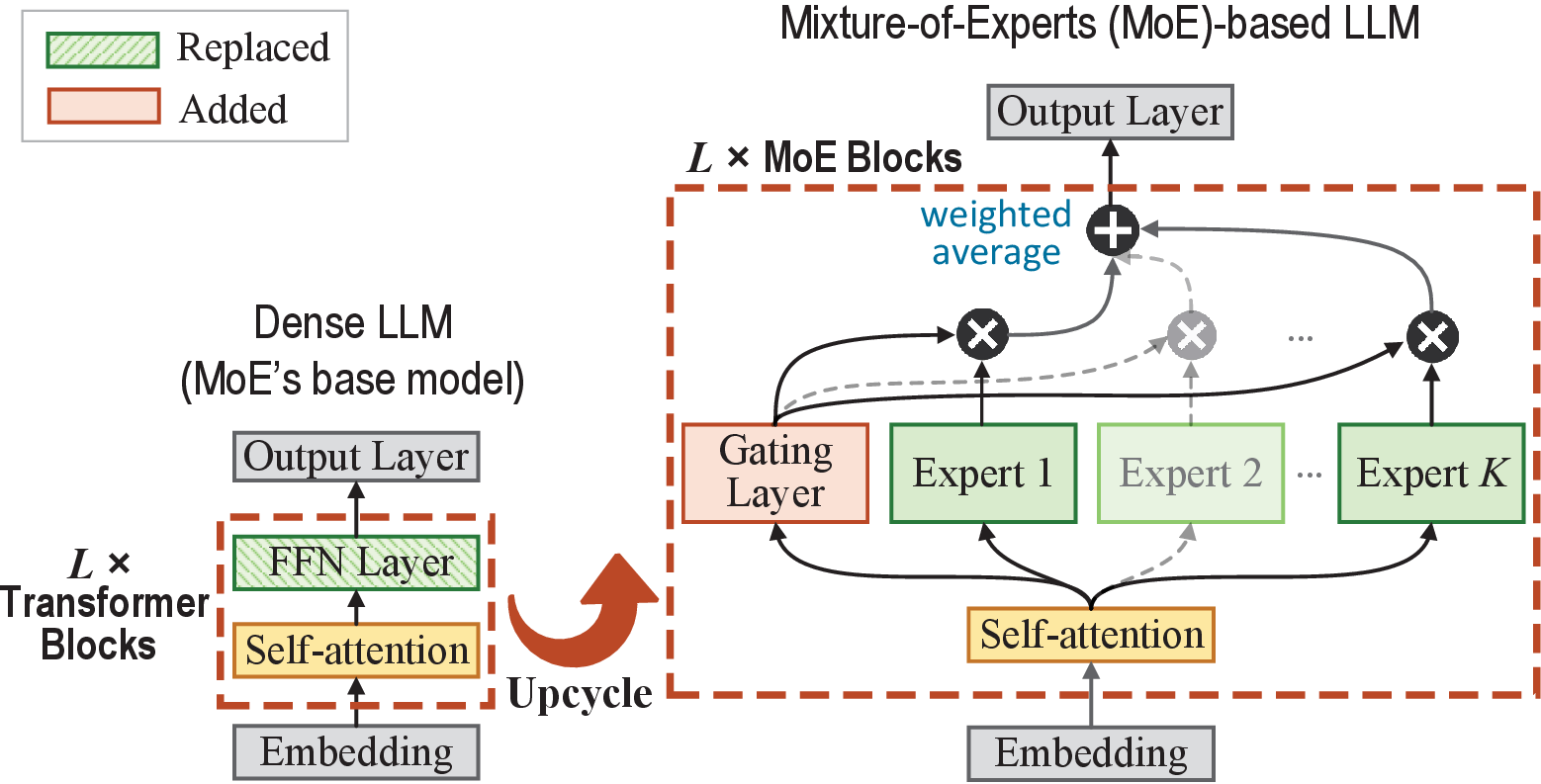}
\caption{Upcycling a dense LLM to an MoE-based LLM.}
\label{fig:moe_model}
\end{figure}

\section{Preliminaries}\label{sec:preliminaries}
\textbf{Autoregressive Language Modeling in LLMs:} The LLMs are typically trained with autoregressive language modeling \cite{Vaswani2017Attention} to generate coherent and contextually appropriate text. Given an input sequence of tokens (words or subwords), the LLM predicts the most likely next token and and appends it to the input sequence for subsequent token predictions. This process repeats iteratively until the LLM generates a special end-of-sentence token (e.g., \small{$<$}\normalsize eos\small{$>$}\normalsize), signaling the completion of text generation.

Let $\mathbf{x}=\{x_0, x_1, ..., x_{T}\}$ define a reference sequence of tokens where $x_t$ denotes the $t$-th token in the sequence. The LLM training objective is to maximize the likelihood of predicting the correct token $x_t$, given the previous tokens $x_0, x_1, ..., x_{t-1}$ for all $t\in\{1,...,T\}$:
\begin{equation}\label{eq:p_llm}
P_{\mathrm{LLM}}(\mathbf{x} ; m)=\prod_{t=1}^T P\left(x_t \mid x_1, ..., x_{t-1} ; m\right)\textup{,}
\end{equation}
where $P\left(x_t \mid x_1, \ldots, x_{t-1} ; m\right)$ denotes the conditional probability of predicting the correct token $x_t$ given the previous tokens $x_1, \ldots, x_{t-1}$ and LLM parameters $m$. In practice, autoregressive LLMs evaluate the cross-entropy loss $\boldsymbol{L}_{\mathrm{CE}}(\mathbf{x} ; m)$ for the reference token sequence $\mathbf{x}$ by computing the negative log-likelihood of the predicted tokens. The LLM parameters $m$ are optimized using SGD-like algorithms to minimize the training loss.
\begin{equation}\label{eq:ce_loss}
\boldsymbol{L}_{\mathrm{CE}}(\mathbf{x}; m)\! =\! -\!\log P_{\mathrm{LLM}}(\mathbf{x}; m)\textit{.}
\end{equation}

As training progresses, the LLM will be increasingly adept at the reference NLP context and more confident in token predictions. Token perplexity $\Gamma(\mathbf{x} ; m)$ \cite{Xia2023Scaling} is a common metric for assessing the LLM performance, which quantifies the average uncertainty in predicting the correct token:
\begin{equation}
\Gamma(\mathbf{x} ; m)=\left(\prod_{t=1}^T \frac{1}{P\left(x_t \mid x_1, \ldots x_{t-1} ; m\right)}\right)^{1 / T}.
\end{equation}
A low token perplexity $\Gamma(\mathbf{x} ; m)$ indicates that the LLM has adapted well to the reference NLP task, consistently making correct token predictions. In contrast, a high token perplexity $\Gamma(\mathbf{x} ; m)$ suggests that the LLM remains uncertain or frequently makes incorrect token predictions.
% Ideally, if the LLM is perfectly confident and always predicts tokens accurately, the optimal token perplexity is 1.

\textbf{Mixture-of-Experts (MoE) Architecture:} The MoE-based LLMs, such as Qwen-MoE \cite{Qwen_MoE} and DeepSeek-MoE \cite{DeepSeek_MoE}, have emerged as a growing trend of the mainstream LLMs. As shown in Fig. \ref{fig:moe_model}, the MoE models are typically upcycled from traditional dense LLMs. The dense LLM is built from a sequential stack of $L$ transformer blocks, where each transformer block contains a multi-head self-attention module and a feed-forward network (FFN). The MoE model retains its backbone architecture, but upgrades each transformer block into an MoE block by replacing the FFN with $K$ specialized, FFN-based experts \cite{Lepikhin2021GShard}.

Each MoE block contains experts that capture different aspects of domain knowledge to process input tokens. Unlike dense LLMs that activate all model parameters for an input token, the MoE models employ a sparsely activation strategy. For each input token, an MoE block dynamically activates a subset of experts aligned with the most relevant knowledge domains for processing.

The activation of experts is controlled by a gating mechanism, which operates as a softmax classifier (shown as the gating layer in Fig. \ref{fig:moe_model}). For a given input token, the gating layer calculates the probability of each expert being selected to process that token. Meanwhile, a parameter $k$, where $k < K$, specifies that only the experts with top-$k$ selection probabilities are activated. The computation outputs of those activated experts are combined and weighted averaged to derive the final output of the current MoE block. Formally, the final computation result $y$ of the MoE block for an input $x$ is:
\begin{equation}
y = \sum\nolimits_{i\,\in\,\text{top}_k} p_i \cdot y_i(x)
\end{equation}
where $p_i$ denotes the selection probability of expert $i$, and $y_i(x)$ is the output of expert $i$ for input $x$. The advanced MoE architecture enables itself to capture diverse domain knowledge across different experts, thus learning more complex feature representations and enabling superior performance over traditional dense LLMs.

\begin{figure}[t]
\centering
\includegraphics[width= 3.5in, keepaspectratio]{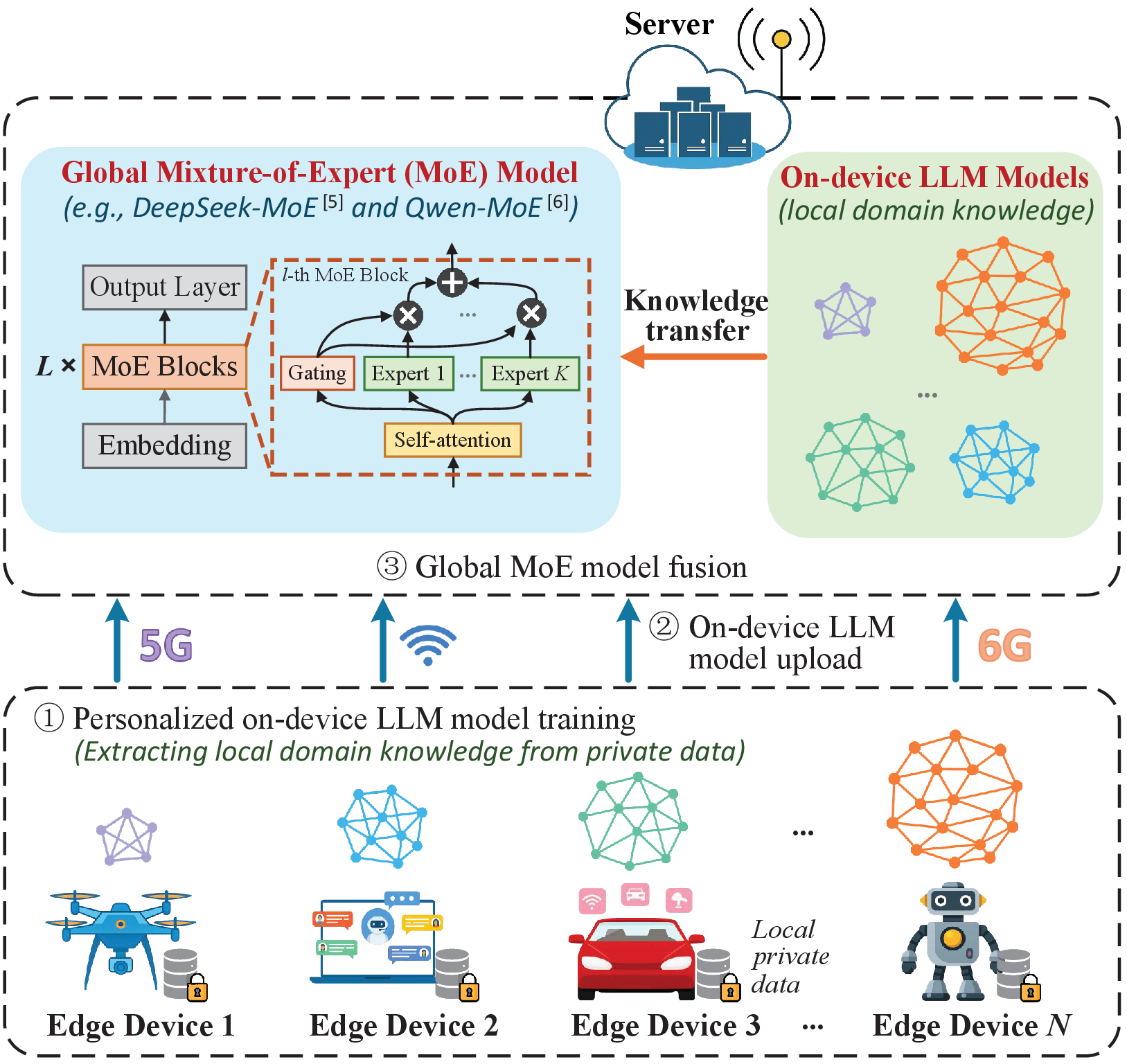}
\caption{{\sc DeepFusion} system: Federated MoE training via knowledge transfer from heterogeneous edge devices.}
\label{fig:system_model}
\end{figure}

\begin{figure*}[t]
\centering
\includegraphics[width= 7.15in, keepaspectratio]{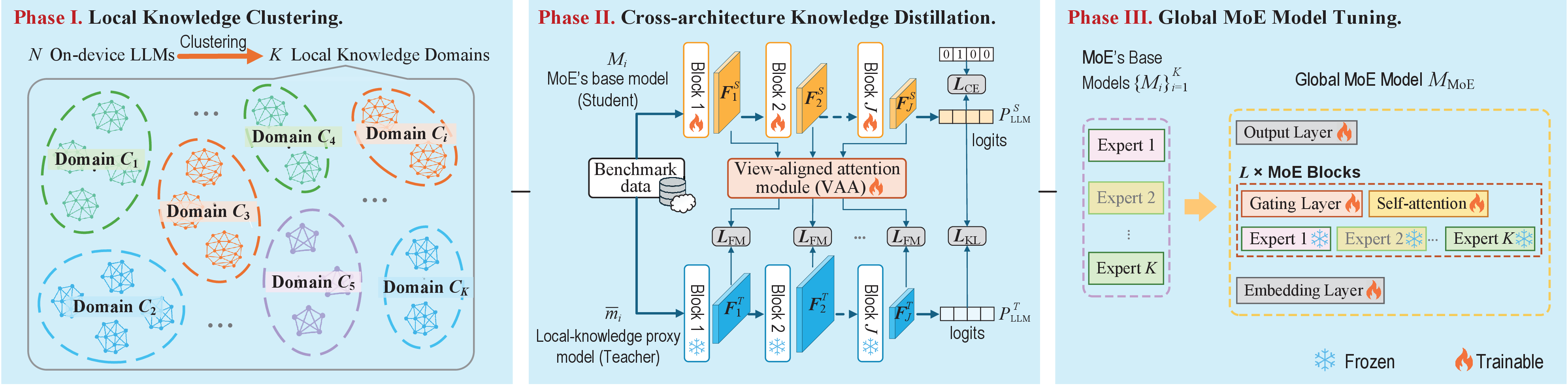}
\caption{Server-side pipeline for federated knowledge distillation from heterogeneous on-device LLMs to the global MoE model.}
\label{fig:algorithm_workflow}
\end{figure*}

\section{System Overview}\label{sec:system_model}
\subsection{System Description}
As shown in Fig. \ref{fig:system_model}, we envision a federated MoE learning system across $N$ heterogeneous edge devices in set $\mathcal{N}$. The goal is to collaboratively train a global MoE-based LLM for natural language generation tasks. Each edge device $n\in\mathcal{N}$, representing ubiquitous AIoT devices in real world, runs its on-device LLM model $m_n$ for local AI applications. To best meet its local application requirements and resource constraints, each edge device independently tailors the architecture and parameter size of its local LLM model $m_n$. All edge devices generate privacy-sensitive local application data, which strictly remains on-device.

A powerful central server coordinates the federated MoE learning process, aggregating knowledge from each on-device LLM model into the global MoE model. Each on-device model specializes as a domain-specific expert by training on local application data for its specific AI scenario. The expertise from heterogeneous on-device LLMs serves as a valuable knowledge source for the multiple experts within the MoE. Through knowledge distillation, the central server efficiently integrates the expertise of on-device LLMs into the global MoE model. The resulting MoE-based LLM consolidates domain-specific knowledge from all devices, enabling robust performance on a wide range of natural language queries. To summarize, our objective is to leverage distributed expertise from heterogeneous on-device LLMs $\{m_n\}_{n=1}^N$ to construct a robust, knowledge-abundant global MoE model $M_{\mathrm{MoE}}$.

\subsection{Key Challenges of System Implementation}\label{subsec:challenges}
It poses significant technical challenges to realize a federated MoE training system in practical edge environments, specifically in these areas:
\begin{itemize}[leftmargin=*]
\item \textbf{Challenge 1 (FL Communication Costs):} Traditional FL requires multiple communication rounds to exchange local model updates between edge devices and the central server. However, modern on-device LLMs, such as GPT-2 Medium \cite{Radford2019Language} (380M parameters) and Tiny-LLaMa \cite{Zhang2024Tiny} (1.1B parameters), are increasingly significant, often exceeding hundreds of millions to over a billion parameters. The growing size of on-device LLMs results in high FL communication costs over resource-constrained edge networks.
\item \textbf{Challenge 2 (System Scalability):} An increasing number of participating edge devices, which communicate with the central server, would exacerbate FL communication overhead on resource-constrained edge networks. Each on-device LLM captures unique local domain knowledge that is transferable to the global MoE model. Nonetheless, performing knowledge distillation for each on-device model would impose significant computational overhead on the central server.

\item \textbf{Challenge 3 (Edge Device Heterogeneity):} The on-device LLMs (teacher models) are typically based on compact LLM architectures \cite{Zheng2025Review}, differing from the global MoE-based LLM (student model) in model architecture. As a result, the teacher and student models reside in distinct latent spaces. Traditional federated knowledge distillation methods, relying on direct teacher-student logits or feature alignment in a shared latent space, are infeasible for cross-architecture knowledge transfer.
\end{itemize}

Motivated by the aforementioned challenges, we propose a comprehensive solution framework named {\sc DeepFusion}. The methodology details are presented in Section~\ref{sec:methodology}.

\section{Methodology}\label{sec:methodology}
\textsc{DeepFusion} aims to enable communication-efficient and scalable federated knowledge transfer from heterogeneous edge devices $n \in \{1, \ldots, N\}$, ultimately obtaining a global MoE model $M_{\mathrm{MoE}}$ that ensembles the distributed knowledge of on-device LLMs $\{m_n\}_{n=1}^N$. As shown in Fig. \ref{fig:algorithm_workflow}, federated knowledge distillation workloads are computation-intensive, thus they are handled by the central server. In contrast, each edge device $n$ simply trains its lightweight on-device LLM $m_n$ on private data to acquire local domain knowledge. This enables broad participation especially from resource-limited edge devices, aligning with our goal to sourcing diverse and extensive training data from ubiquitous edge devices for MoE training. In the following, we elaborate on the design principles and key modules of \textsc{DeepFusion}.

\subsection{One-shot Federated Learning Design}
In contrast to traditional FL with multiple communication rounds, \textsc{DeepFusion} employs a one-shot FL design to enhance communication efficiency. Specifically, each edge device $n$ trains its on-device LLM $m_n$ using private local application data. As local training advances, the on-device LLM's performance is consistently optimized, while progressively capturing essential local domain knowledge. Once local training is complete, the fully optimized on-device LLM $m_n$ is uploaded to the central server in \textit{a single communication round} for federated knowledge distillation. Thus, the total communication cost $\digamma^{\textup{net}}$ of \textsc{DeepFusion} is:
\begin{equation}
\digamma^{\textup{net}} = \sum\nolimits_{n=1}^N \left|m_n\right|\textup{,}
\end{equation}
where $\left|m_n\right|$ denotes the data size of on-device LLM $m_n$ to be transmitted to the server.

This one-shot FL design is both efficient and practical. Edge devices are self-motivated to optimize their on-device LLMs, thus enhancing their local AI applications. The central server then acts as a free rider, distilling local domain knowledge from these pre-trained on-device LLMs into the global MoE model $M_{\mathrm{MoE}}$. This process imposes no additional computational or resource burden on the edge devices and improves the utility of local LLM training.

\subsection{Local Knowledge Clustering}
As the number of participating edge devices increases, performing knowledge distillation individually from each on-device LLM into the MoE model $M_{\mathrm{MoE}}$ would impose an unsustainable computational burden on the central server. To tackle this scalability challenge, we propose a local knowledge clustering method. As shown in Phase I of Fig. \ref{fig:algorithm_workflow}, the on-device LLMs uploaded from numerous edge devices are clustered into $K$ \textit{local knowledge domains} $\{C_i\}_{i=1}^K$, matching the $K$ experts in the global MoE model. Subsequently, we construct $K$ \textit{local-knowledge proxy models} $\{\bar{m}_i\}_{i=1}^K$ for these domains $\{C_i\}_{i=1}^K$, upon which we perform knowledge distillation.

\textbf{Implementation Details:} Each local knowledge domain $C_i$ consists of on-device LLM models of the same type that captures similar aspects of local knowledge (e.g., medical expertise on a particular disease). The local knowledge similarity between on-device LLMs is assessed using meta-information from their local training data. When an edge device $n$ uploads the pre-trained on-device LLM $m_n$, it also sends low-rank feature embeddings $e_n$ computed from its raw training data. These feature embeddings $e_n$ capture the semantic information of local data while preserving privacy, indicating the local knowledge domain learned by the on-device LLM $m_n$. Modern on-device NLP semantic encoders (e.g., MiniLM \cite{Wang2021MiniLM}) can efficiently compute these low-rank feature embeddings, which are typically tens of bytes and incur negligible communication costs when sent to the server.

Formally, let $\Pi=[\pi_{n_1, n_2}]\in\mathbb{R}^{N\times N}$ represent the local knowledge similarity matrix, where each element $\pi_{n_1, n_2}$ denotes the cosine similarity between the feature embeddings of on-device LLMs $m_{n_1}$ and $m_{n_2}$:
\begin{equation}
\pi_{n_1, n_2} = \frac{e_{n_1}\cdot e_{n_2}}{\|e_{n_1}\|\cdot \|e_{n_2}\|}\textup{.}
\end{equation}
Based on this similarity matrix $\Pi$, the on-device LLMs are grouped into $K$ local knowledge domains $\{C_i\}_{i=1}^K$ using KMeans.

\begin{figure}[t]
\centering
\includegraphics[width= 2.78in, keepaspectratio]{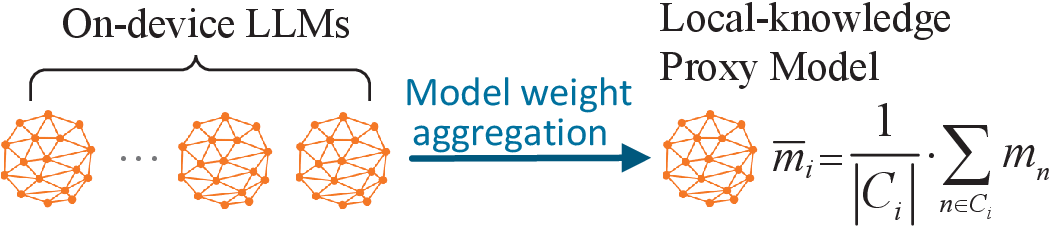}
\caption{Generating a proxy model $\bar{m}_i$ for the local knowledge domain $C_i$.}
\label{fig:local_knowledge_proxy_model}
\end{figure}

Then, as depicted in Fig. \ref{fig:local_knowledge_proxy_model}, within each local knowledge domain $C_i$, a proxy model $\bar{m}_i$ is generated by averaging the weights of the clustered on-device LLMs $m_i \in C_i$. Each proxy model $\bar{m}_i$ encapsulates its respective domain knowledge for transfer into the global MoE model $M_{\mathrm{MoE}}$. There are totally $K$ local-knowledge proxy models, and each proxy model $\bar{m}_i$ is to be distilled into one of $K$ experts in $M_{\mathrm{MoE}}$. In summary, our local knowledge clustering approach effectively enables scalable transfer of diverse local domain knowledge from numerous edge devices into the global MoE model.

\subsection{Cross-architecture Knowledge Distillation}
We will distill the obtained $K$ local-knowledge proxy models $\{\bar{m}_i\}_{i=1}^K$ into the $K$ experts of the global MoE model $M_{\mathrm{MoE}}$, with each proxy model assigned to a corresponding expert. As noted in Section \ref{sec:preliminaries}, an MoE-based LLM model is typically upcycled from a base LLM model. Building on this idea, we first distill each local-knowledge proxy model $\bar{m}_i$ into a separate MoE's base model $M_i$, yielding a collection of $K$ MoE's base models $\{M_i\}_{i=1}^K$. Following the efforts of~\cite{Xia2024AeroReC, Lin2020Ensemble}, we assume that public benchmark data such as from Hugging Face or GitHub is available at the server to facilitate this distillation process. Similar to the model upcycling process, these $K$ MoE's base models are then merged and upgraded into a global MoE model by assigning each base model's parameters to the corresponding expert position in the MoE. The resulting global MoE model $M_{\mathrm{MoE}}$ integrates diverse domain knowledge and can therefore effectively handles diverse input tasks.

When distilling the local-knowledge proxy model into the MoE's base model, a key challenge arises from the discrepancy in predictive perspectives between the local-knowledge proxy model $\bar{m}_i$ (teacher) and the MoE's base model $M_i$ (student). This stems from their distinct model architectures. The local-knowledge proxy model $\bar{m}_i$, constructed by weight-averaging a cluster of on-device LLMs, thus adopts a compact LLM architecture with a small parameter count. This compact LLM architecture progressively builds global semantic features from simpler and localized linguistic cues, achieving global coherence mainly in the final layers \cite{Zheng2025Review}. In contrast, the MoE's base model $M_i$ typically employs a standard LLM architecture with a larger parameter count, which enables it to easily learn global semantic features from the outset \cite{Wang2024Beyond}. This difference in how the teacher and student models learn feature representations leads to a misalignment in their predictive perspectives. Hence, traditional knowledge distillation approaches based on direct teacher-student logits or feature matching~\cite{Eldar2022Federated} are ineffective in our scenario. These methods typically require the teacher and student to have similar model architectures and aligned predictive perspectives, which does not hold in our scenario.

\begin{figure}[t]
\centering
\includegraphics[width= 1.35in, keepaspectratio]{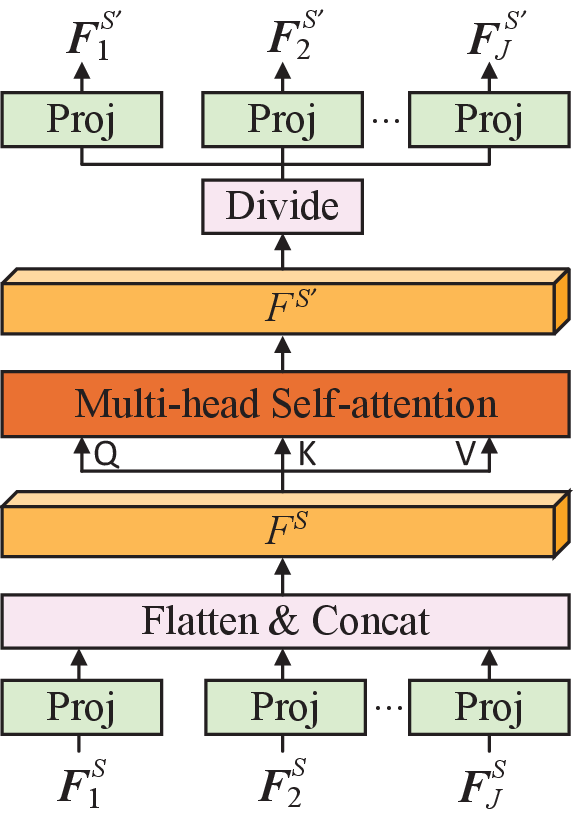}
\caption{Design of the View-aligned Attention (VAA) module.}
\label{fig:vaa}
\end{figure}

To tackle this problem, we design a View-Aligned Attention (VAA) module to support cross-architecture knowledge distillation. As shown in Phase II of Fig. \ref{fig:algorithm_workflow}, both the local-knowledge proxy model $\bar{m}_i$ (teacher) and the MoE's base model $M_i$ (student) are split into $J$ representation stages (blocks). The proposed VAA module employs a multi-head attention mechanism, enabling the student model to integrate its features $\{F_j^S\}_{j=1}^J$ from various representation stages 1 to $J$, thus acquiring a predictive perspective $\{F_j^{S^{\prime}}\}_{j=1}^J$ comparable to that of the teacher model. Owing to the flexibility of the multi-head attention module, the VAA module can be incorporated into any model architecture without requiring specialized design. Specifically, the proposed VAA module is illustrated in Fig. \ref{fig:vaa} and operates in the following three steps:

\textit{First}, we patchify the student features at each representation stage $j\in\{1,\ldots,J\}$ into $\sfrac{P_q}{J}$ patches, capturing detailed feature at each stage. These patches are then projected via convolutional layers $C_j$ to a predefined dimension $\mathbb{R}^{\frac{P_q}{J} \times d}$, where $P_q$ is a hyper-parameter denoting the total number of queries across all $J$ representation stages. These resulting features from all stages are then concatenated to form $F^S \in \mathbb{R}^{P_q \times d}$:
\begin{equation}
F^S=\operatorname{Concat}\left(C_1\left(F_1^S\right), . ., C_J\left(F_J^S\right)\right)\textup{.}
\end{equation}

\textit{Second}, we apply a multi-head self-attention module to $F^S$:
\begin{equation}
F^{S^{\prime}}=\operatorname{Softmax}\left(\frac{\left(W_q F^S\right)\cdot\left(W_k F^S\right)^T}{\sqrt{d}}\right)\cdot W_v \cdot F^S\textup{,}
\end{equation}
where $W_q$, $W_k$, and $W_v$ are trainable self-attention weights, and $d$ denotes the self-attention channel dimension. $F_S$ contains the features from different representation stages of the MoE's base model $M_i$ (student). Through our VAA attention mechanism, the student model learns to focus on different representation stages to blend a new student feature $F^{S^{\prime}}$ whose predictive perspective closely aligns with that of the local-knowledge proxy model $\bar{m}_i$ (teacher).

\textit{Finally}, we divide the blended student features $F^{S'}$ back into $J$ representation stages and project each to match the corresponding stage feature size of the teacher model $\bar{m}i$, resulting in $\{F_j^{S^{\prime}}\}_{j=1}^J$.  As the blended student features $\{F_j^{S^{\prime}}\}_{j=1}^J$ approximate the teacher model's perspective, standard feature alignment can be subsequently  applied:
\begin{equation}
\boldsymbol{L}_{\mathrm{FM}} = \sum\nolimits_{j=1}^J \mathrm{MSE}\left(F_j^T, F_j^{S^{\prime}}\right)
\end{equation}
where $\boldsymbol{L}_{\mathrm{FM}}$ denotes the feature-matching loss computed between the features of the teacher and student models across the $J$ representation stages. Meanwhile, the logits-based distillation loss $\boldsymbol{L}_{\mathrm{KL}}$ using KL-divergence functions is defined:
\begin{equation}
\boldsymbol{L}_{\mathrm{KL}} = \mathrm{KL}\left(P_{\mathrm{LLM}}^T, P_{\mathrm{LLM}}^S\right)
\end{equation}
where $P_{\mathrm{LLM}}^T$ and $P_{\mathrm{LLM}}^S$ indicate the soft-label predictions (logits) of teacher and student models, as computed in Eq. (\ref{eq:p_llm}). The loss $\boldsymbol{L}_{\mathrm{KL}}$ minimizes the discrepancy between the teacher's and student's final logits. The cross-entropy loss $\boldsymbol{L}_{\mathrm{CE}}$ for the student model $M_i$ on final hard-label predictions is described in Eq. (\ref{eq:ce_loss}). To sum up, the total cross-architecture knowledge distillation loss $\boldsymbol{L}_{\mathrm{KD}}$ between $\bar{m}_i$ and $M_i$ is:
\begin{equation}
\boldsymbol{L}_{\mathrm{KD}} = \boldsymbol{L}_{\mathrm{CE}} + \alpha\cdot\boldsymbol{L}_{\mathrm{FM}} + \beta\cdot\boldsymbol{L}_{\mathrm{KL}}
\end{equation}
where $\alpha$ and $\beta$ are trade-off hyper-parameters controlling the impact of each loss term. The MoE's base model $M_i$ (student) is trained to minimize the total cross-architecture knowledge distillation loss $\boldsymbol{L}_{\mathrm{KD}}$, thereby effectively distilling knowledge from the local-knowledge proxy model $\bar{m}_i$ (teacher).

\subsection{Global MoE Model Tuning}
\begin{figure}[t]
\centering
\includegraphics[width= 3.5in, keepaspectratio]{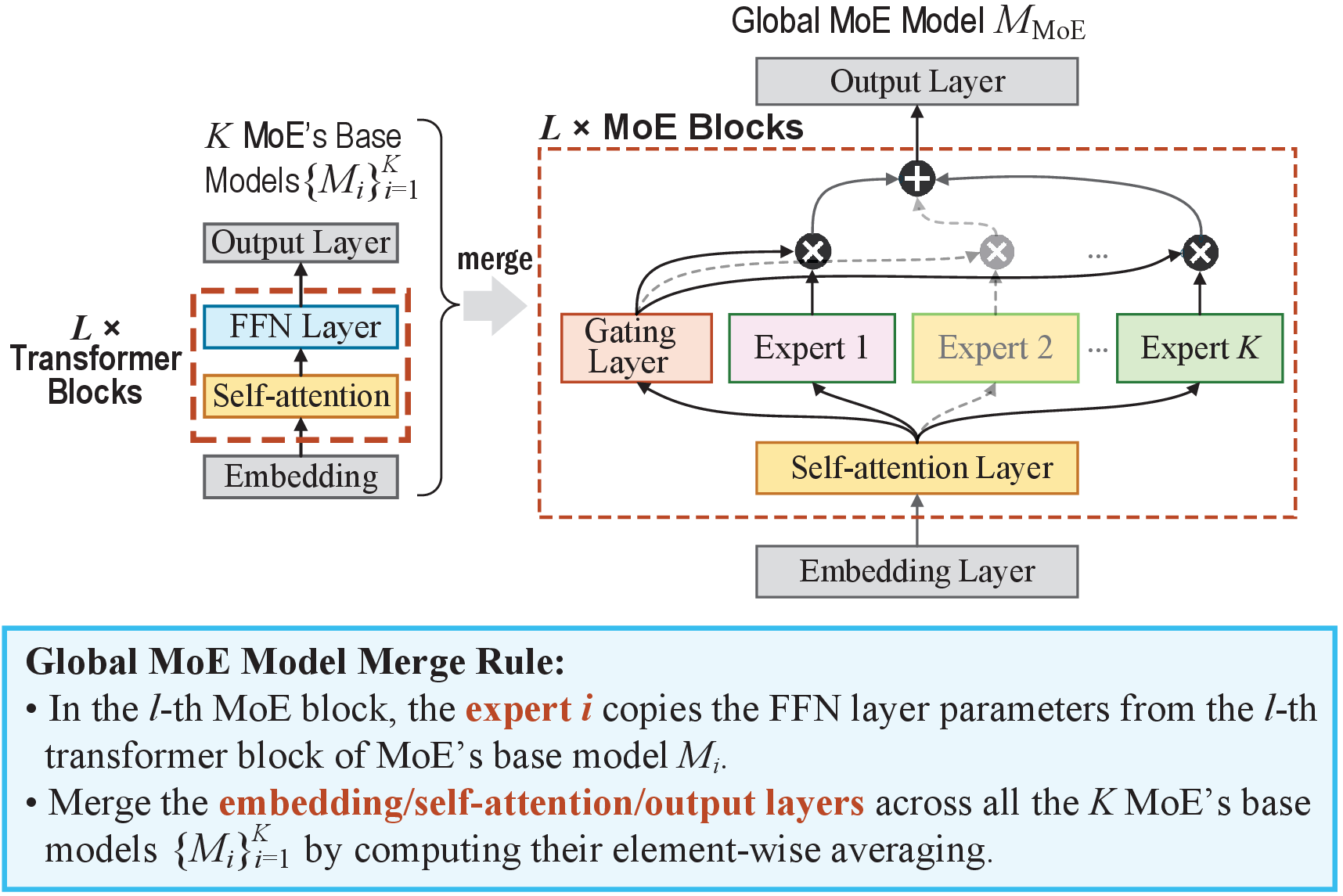}
\caption{Merging the $K$ MoE's base models $\{M_i\}_{i=1}^K$ into the global MoE model $M_{\mathrm{MoE}}$.}
\label{fig:moe_merge}
\end{figure}

As in Phase III of Fig. \ref{fig:algorithm_workflow}, we will merge the obtained $K$ MoE's base models $\{M_i\}_{i=1}^K$, each capturing unique domain knowledge from heterogeneous edge devices, into the global MoE model $M_{\mathrm{MoE}}$. Fig. \ref{fig:moe_merge} depicts the global MoE model merge rule. In the global MoE model $M_{\mathrm{MoE}}$, each expert $i$, which is built using FFN layers, directly inherits the parameters of the FFN layers $M_i^{\text{FFN}}$ from its corresponding base model $M_i$:
\begin{equation}
M^{i}_{\text{expert}} \leftarrow M_i^{\text{FFN}}\text{.}
\end{equation}
This leverages the fact that the FFN layers in different MoE's base models $\{M_i\}_{i=1}^K$ capture distinct domain-specific knowledge for processing input tokens, enabling the FFN-based experts in $M_{\mathrm{MoE}}$ to specialize accordingly. In contrast, the layers such as the embedding, self-attention, and output layers in $M_{\mathrm{MoE}}$ operate globally to capture relationships and dependencies between tokens in a sequence. Therefore, we initialize these layers by averaging the parameters of the corresponding layers from the $K$ MoE's base models ${M_i}_{i=1}^K$, thus aggregating the diverse knowledge encoded in these components across different MoE's base models:
\begin{equation}
M_{\mathrm{MoE}}^x=\frac{1}{K}\cdot\sum\nolimits_{i=1}^K M_i^x\text{,}
\end{equation}
where $x$ refer to the embedding, self-attention or output layers.

After initializing the global MoE model $M_{\mathrm{MoE}}$ with the $K$ base models $\{M_i\}_{i=1}^K$, we start with global MoE model tuning. The FFN-based experts are frozen, while the other layers including embedding, self-attention, gate, and output layers are fine-tuned. Based on public benchmark data at the server, the gate layer $M_{\mathrm{MoE}}^{gate}$ is trained to compute optimal expert selection probabilities $\langle p_i \rangle_{i=1}^K$ for each input token, enabling sparse activation of the most relevant top-$k$ experts. The embedding, self-attention, and output layers are jointly trained to more effectively capture global relationships and dependencies between tokens in a sequence. Finally, the global MoE model $M_{\mathrm{MoE}}$ converges to optimal performance.

\captionsetup{skip=1pt}
\begin{table*}[t]
\centering
\renewcommand\arraystretch{1.5}
\small
\begin{threeparttable}
\caption{Token Perplexity ($\log$) Results Under Different Case Studies and Various System Scale Settings.}
\begin{tabularx}{\textwidth}{l|X|X|X|X|X|X|X|X}
\hline
\multirow{2}{*}{\textbf{Method}}
  & \multicolumn{4}{c|}{\textbf{(1) Qwen-MoE for Medical Multi-choice QA}}
  & \multicolumn{4}{c}{\textbf{(2) DeepSeek-MoE for Financial Open-ended QA}} \\
  & $N=16$ & $N=32$ & $N=64$ & $N=128$ & $N=16$ & $N=32$ & $N=64$ & $N=128$ \\
\hline
FedJETS \cite{Dun2023Fedjets}    & 0.1639 & 0.1588 & 0.1557 & 0.1571 & \uwave{3.9717} & 3.9863 & 3.9987 & \uwave{4.0006} \\
FedKMT \cite{Fan2025FedMKT}   & 0.1609 & 0.1574 & 0.1526 & 0.1546 & 3.9908 & \uwave{3.9811} & \uwave{3.9941} & 4.0030 \\
OFA-KD \cite{Hao2023OFA}   & \textbf{0.1573} & \textbf{0.1535} & \textbf{0.1510} & \textbf{0.1494} & 4.0375 & 4.0398 & 4.0224 & 4.0333 \\
\hline
\textbf{\textsc{DeepFusion}} & \uwave{0.1603} & \uwave{0.1569} & \uwave{0.1551} & \textbf{0.1494} & \textbf{3.9697} & \textbf{3.9700} & \textbf{3.9774} & \textbf{3.9723} \\
\hline
\end{tabularx}
\label{tab:perplexity_comparison}
\end{threeparttable}

\end{table*}

\captionsetup{skip=1pt}
\begin{table*}[t]
\centering
\renewcommand\arraystretch{1.5}
\small
\begin{threeparttable}
\caption{Token Accuracy ($\%$) Results Under Different Case Studies and Various System Scale Settings.}
\begin{tabularx}{\textwidth}{l|X|X|X|X|X|X|X|X}
\hline
\multirow{2}{*}{\textbf{Method}}
  & \multicolumn{4}{c|}{\textbf{(1) Qwen-MoE for Medical Multi-choice QA}}
  & \multicolumn{4}{c}{\textbf{(2) DeepSeek-MoE for Financial Open-ended QA}} \\
  & $N=16$ & $N=32$ & $N=64$ & $N=128$ & $N=16$ & $N=32$ & $N=64$ & $N=128$ \\
\hline
FedJETS \cite{Dun2023Fedjets}   & \uwave{92.45} & \text92.39 & 92.36 & 92.41 & \uwave{28.28} & 29.47 & 29.03 & 28.64 \\
FedKMT \cite{Fan2025FedMKT}    & 92.34 & \textbf{92.42} & \textbf{92.42} & 92.38 & 28.26 & \uwave{29.57} & \textbf{29.06} & 28.93 \\
OFA-KD \cite{Hao2023OFA}  & 92.38 & 92.38 & 92.41 & \textbf{92.55} & 27.89 & 28.88 & 29.04 & \uwave{28.95} \\
\hline
\textbf{\textsc{DeepFusion}} & \textbf{92.52} & \uwave{92.41} & \textbf{92.42} & \uwave{92.45} & \textbf{28.32} & \textbf{29.67} & \textbf{29.06} & \textbf{29.22} \\
\hline
\end{tabularx}
\begin{tablenotes}
\small
\item * We \textbf{bold} the highest performance and \uwave{underline} the second highest performance for each row.
\end{tablenotes}
\label{tab:accuracy_comparison}
\end{threeparttable}
\end{table*}

Since only the embedding, self-attention, gate, and output layers are fine-tuned while the larger portion of FFN-based experts remain frozen, the computational overhead is minimal. The FFN-based experts constitute the majority of model parameters. Their weights have already been determined by leveraging heterogeneous federated edge devices to contribute local domain knowledge distilled from their pre-trained on-device LLMs. These trainable components represent only a small fraction of total model parameters, resulting in reduced memory footprint and faster optimal model convergence. Our approach distinguishes from traditional centralized MoE training at cloud servers \cite{Rajbhandari2022Deepspeed, Hu2025Communication}, which typically requires extensive full-parameter training.

\section{Performance Evaluation}\label{sec:evaluation}
\subsection{Experimental Setup}
\textbf{Case Studies:} To evaluate \textsc{DeepFusion}'s generalizability, we conduct two realistic case studies: (1) medical multi-choice question answering (QA) and (2) financial open-ended QA. Each case involves training an industry-scale global MoE-based LLM, specifically Qwen1.5-MoE (14.3B Param.) and DeepSeek-MoE-16B-base.

\begin{comment}
\captionsetup{skip=3.5pt}
\begin{table}[t]
\centering
\caption{RAM Requirement for Training On-device LLMs}
\vspace{-0.5ex}
\renewcommand\arraystretch{1.2}
\small
\begin{tabular}{l|c}
\toprule[1.25pt]
\textbf{Model}             & \textbf{Runtime Memory}  \\ \hline
GPT-2 (137M Param.)        &   2.58 GB                              \\ \hline
GPT-2 Medium (380M Param.) &  3.73 GB                                \\ \hline
TinyLlama (1.1B Param.)    &  7.25 GB                            \\ \hline
OLMo (1.2B Param.)         & 5.61 GB                            \\ \hline
BLOOM (1.1B Param.)  &   4.80 GB                              \\ \bottomrule[1.25pt]
\end{tabular}
\label{tab:on_device_llms}
\vspace{-4.5ex}
\end{table}
\end{comment}

\noindent\textbf{(1) Qwen-MoE for medical multi-choice QA:} We use the MMedBench dataset \cite{Qiu2024MMedBench}, a multilingual medical multi-choice dataset covering six languages. Each input sample consists of a medical question and its corresponding candidate answers. The global Qwen1.5-MoE is trained to select the correct answer for each question. We split 7,669 data samples for the testset, and 30,429 samples for local training data over edge devices. The on-device LLMs deployed on edge devices include GPT-2, GPT-2 Medium and TinyLlama.

\noindent\textbf{(2) DeepSeek-MoE for financial open-ended QA:} We use the FinQA dataset \cite{fingpt_fiqa_qa}, a financial open-ended QA dataset. Each input sample consists of an instruction describing the finance QA context and a question that to be answered directly. The global DeepSeek-MoE-16b-base model is trained to generate a paragraph-style answer for each question. Similarly, we split 1,248 samples for the testset, and 10,618 for local training data over edge devices. The on-device LLMs involve TinyLlama, OLMo-1.2B and BLOOM-1.1B.

\begin{figure*}[t] % use figure* for full page width if in two-column
    \centering
    % Fig. 7: left (width ratio 1/3)
    \begin{minipage}[t]{0.345\textwidth}
        \centering
        \includegraphics[width=\linewidth, keepaspectratio]{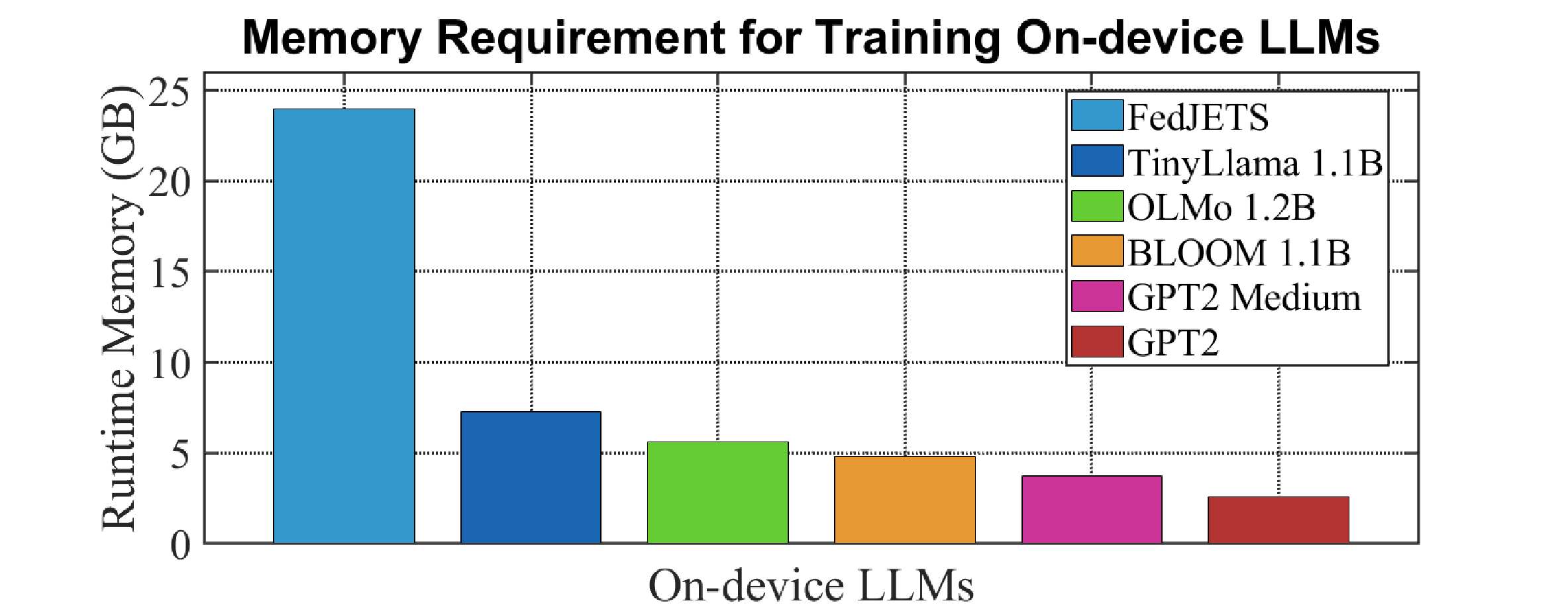}
        \caption{On-device Memory Footprint.}
        \label{fig:ram_memory}
    \end{minipage}%
    % Fig. 8: right (width ratio 2/3)
    \begin{minipage}[t]{0.655\textwidth}
        \centering
        \includegraphics[width=\linewidth, keepaspectratio]{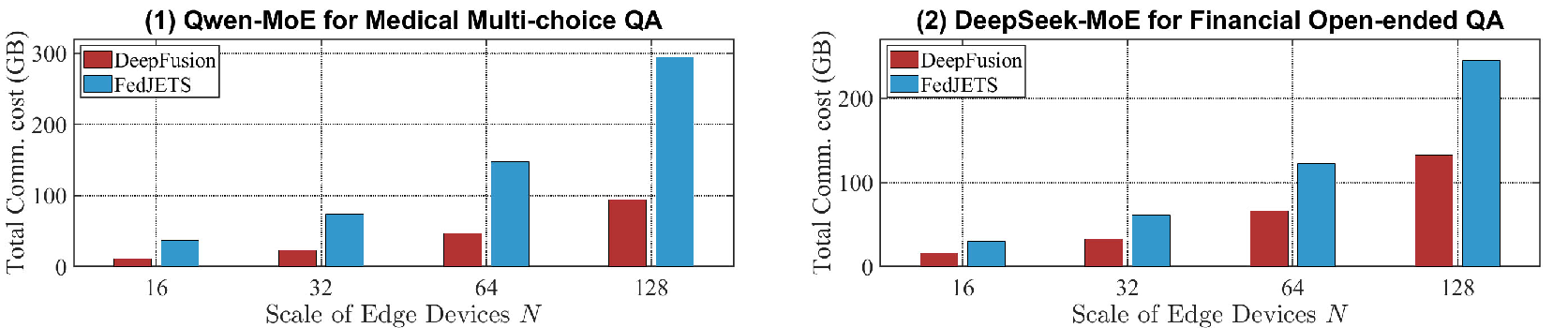}
        \caption{Communication Costs of Different Federated MoE Training Methods.}
        \label{fig:comm_costs}
    \end{minipage}
\end{figure*}

Our selected on-device LLMs encompass various model architectures and sizes, enabling deployment on a range of edge devices with varying hardware capacities. This supports our considered system scenario, where ubiquitous edge devices participate in federated knowledge distillation-based MoE training. Fig. \ref{fig:ram_memory} depicts the RAM requirements for training these on-device LLMs after bfloat16 and NF4 quantization, which are generally affordable for typical edge devices (e.g., NVIDIA Jetson Nano with 4GB RAM and iPhone 16 with 8GB RAM).

\textbf{System Settings:} We evaluate the performance of \textsc{DeepFusion} by varying the number of participating edge devices from 16, 32, 64 to 128. Each edge device is randomly assigned to operate one of the on-device LLMs used in our case studies. The local private training data is distributed randomly and unevenly across the edge devices.

\begin{comment}
\textbf{Baselines:} We compare \textsc{DeepFusion} against three state-of-the-art methods, including FedJETS \cite{Dun2023Fedjets}, FedKMT \cite{Fan2025FedMKT}, and OFA-KD \cite{Hao2023OFA}:

\noindent\textbf{(1)} FedJETS \cite{Dun2023Fedjets} implements a federated MoE training system by training a compact MoE network, pruned from the global MoE model, on each edge device. These local compact MoE networks are merged into a global MoE model at the server.

\noindent\textbf{(2)} FedKMT \cite{Fan2025FedMKT} is a federated knowledge transfer method between small and large language models. It leverages the final logits produced by small LLMs (teachers) to supervise the global large LLM on the server.

\noindent\textbf{(3)} OFA-KD \cite{Hao2023OFA} is a cross-architecture knowledge distillation approach that projects intermediate student features into the logits space and aligns them with the teacher model's final logits. It can be an ablation baseline to evaluate our proposed VAA module in cross-architecture knowledge distillation.
\end{comment}

\textbf{Baselines:} We compare \textsc{DeepFusion} against a theoretical upper-bound (DeepSpeed \cite{Rajbhandari2022Deepspeed}) and three state-of-the-art methods (FedJETS \cite{Dun2023Fedjets}, FedKMT \cite{Fan2025FedMKT}, and OFA-KD \cite{Hao2023OFA}).

\noindent\textbf{(1)} DeepSpeed \cite{Rajbhandari2022Deepspeed} represents the traditional centralized MoE training approach on cloud servers and serves as the theoretical performance upper-bound for \textsc{DeepFusion}.

\noindent\textbf{(2)} FedJETS \cite{Dun2023Fedjets} implements a federated MoE training system by training a compact MoE network, pruned from the global MoE model, on each edge device. These local compact MoE networks are merged into a global MoE model at the server.

\noindent\textbf{(3)} FedKMT \cite{Fan2025FedMKT} is a federated knowledge transfer method between small and large language models. It leverages the final logits produced by small LLMs to supervise the global large LLM on the server.

\noindent\textbf{(4)} OFA-KD \cite{Hao2023OFA} is a cross-architecture knowledge distillation approach that projects intermediate student features into the logits space and aligns them with the teacher model's final logits. It can be \textit{an ablation baseline} to evaluate our proposed cross-architecture knowledge distillation mechanism.

\begin{figure*}[t] % use figure* for full page width if in two-column
\centering
\includegraphics[width= 6.5in, keepaspectratio]{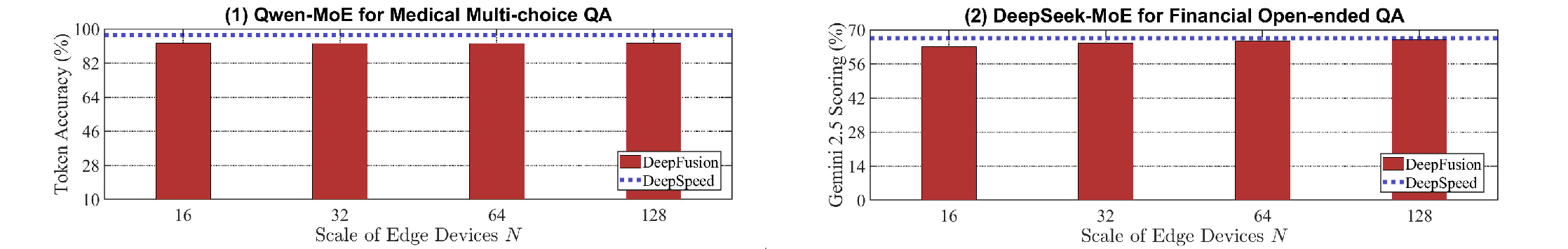}
\caption{Performance Comparison between \textsc{DeepFusion} and DeepSpeed \cite{Rajbhandari2022Deepspeed} (Centralized MoE Training Method).}
\label{fig:deepspeed}
\end{figure*}

\subsection{Main Experimental Results}
\textbf{Performance Comparison:} In Tables~\ref{tab:perplexity_comparison} and~\ref{tab:accuracy_comparison}, we evaluate \textsc{DeepFusion} on two case studies. Specifically, we compare the token perplexity and token accuracy of the resulting global MoE model against three state-of-the-art baselines across different system scales ($N=16$ to $N=128$ edge devices). The token accuracy measures the percentage of tokens in the generated output that exactly match the tokens in the reference sequence. The results show that our \textsc{DeepFusion} effectively distills local domain knowledge from diverse edge devices, enabling the global MoE model to achieve comparable or even superior performance compared to the baselines. Notably, \textsc{DeepFusion} demonstrates robust and superior performance across system scales in the financial open-ended QA case study. Typically, the open-ended QA task is considered more challenging than the multi-choice QA task, as it requires LLMs to possess complex reasoning abilities in order to generate coherent and fluent free-form answers. In this challenging setting, \textsc{DeepFusion} exhibits even greater advantages over the baselines. This is attributed to our designed VAA module which explicitly distills feature knowledge from local models into the global MoE model, and such distilled feature knowledge effectively enhances the reasoning capacity of the resulting MoE-based LLM.

Note that, in the financial open-ended QA task, the token accuracy results do not appear high, as open-ended questions typically allow for diverse and flexible answer expressions, making exact token matches less likely even for semantically correct responses. Nevertheless, token accuracy still reflects how well \textsc{DeepFusion} trains the global MoE model, complementing the token perplexity metric.

\textbf{On-device Memory Footprint:} Fig. \ref{fig:ram_memory} presents the peak memory usage of an edge device during on-device LLM training for our \textsc{DeepFusion} method and the state-of-the-art federated MoE training baseline named FedJETS. \textsc{DeepFusion} enables ubiquitous, resource-limited edge devices to participate in federated MoE training, regardless of local hardware constraints. Each edge device can run lightweight on-device LLMs such as TinyLlama 1.1B, OLMo 1.2B, BLOOM 1.1B, GPT2 Medium, and GPT2, according to its own application needs and resource availability. Once these LLMs are sufficiently trained locally, they serve as repositories of local domain knowledge, transferring knowledge to the global MoE model through knowledge distillation. FedJETS requires deployment of a local MoE network on each edge device. The local expert network, despite its compactness, still retains the MoE architecture and remains too cumbersome for many edge devices. Our empirical measurements indicate that FedJETS demands 3.3x to 9.3x more on-device memory than \textsc{DeepFusion}, which imposes a significant hardware threshold for edge participation. This limits the inclusion of extensive and diverse edge devices in federated MoE training.

\textbf{FL Communication Efficiency:} Fig. \ref{fig:comm_costs} compares the total FL communication costs of DeepFusion and FedJETS across two case studies. \textsc{DeepFusion} allows edge devices to operate and train lightweight on-device LLMs. The on-device LLMs can be those naturally pre-deployed on edge devices for their own application requirements, so \textsc{DeepFusion} does not require additional, dedicated on-device LLM deployment. Furthermore, in a communication-efficient one-shot FL setup, once optimal local performance is achieved, the local LLMs are uploaded to the central server in a single communication round for server-side knowledge distillation into the global MoE model. Conversely, FedJETS assigns each edge device a local expert model, essentially pruned from the global MoE model. These local expert models, usually retaining the complex MoE architecture and typically larger than standard lightweight on-device LLMs, incur significant communication overhead from downloading to edge devices and uploading trained versions back to the server. As the number of edge devices grows, FedJETS' total FL communication costs rise rapidly, while \textsc{DeepFusion} maintains manageable costs across different system scales, as depicted in Figure \ref{fig:comm_costs}.

\subsection{Ablation Analysis}
\textbf{Heterogeneous vs. Homogeneous On-device LLMs:} In Tables~\ref{tab:perplexity_comparison} and~\ref{tab:accuracy_comparison}, our \textsc{DeepFusion} method consistently outperforms FedJETS. This demonstrates the inherent advantages of \textsc{DeepFusion}, which effectively leverages heterogeneous on-device LLMs as the sources of local domain knowledge to be distilled into the global MoE model. Heterogeneous on-device LLMs learn distinct prediction perspectives, shaped by varying model architectures. By distilling the knowledge from these heterogeneous on-device models into the global MoE, \textsc{DeepFusion} enables the aggregation of complementary insights from multiple predictive perspectives, resulting in better and more comprehensive global MoE model performance. This flexibility in supporting heterogeneous on-device LLMs stands in contrast to FedJETS, which require all edge devices to operate a similar compact MoE architecture referred to as a local expert model. In such homogeneous setups, the global MoE model can only integrate knowledge of the same predictive perspectives from homogeneous local expert models, hence limiting the MoE's capacity to generalize and improve.

\textbf{Cross-architecture Knowledge Distillation:} We compare \textsc{DeepFusion} with two cross-architecture knowledge distillation baselines including FedKMT and OFA-KD. Both FedKMT and OFA-KD are logits-driven knowledge distillation approaches, where the logits refer to the soft-label predictions from the final model layer. While lgotis can transfer high-level knowledge, they often neglect implicit model knowledge such as hidden feature representations, which are crucial for effective knowledge distillation. As shown in Tables~\ref{tab:perplexity_comparison} and~\ref{tab:accuracy_comparison}, this limitation is evident in the case study of financial open-ended QA, where the global DeepSeek-MoE model requires advanced knowledge reasoning abilities to generate coherent free-form responses. These reasoning capacities are captured in the teacher models' hidden feature representations, which trace the underlying reasoning process beyond final logits. By employing feature-driven cross-architecture knowledge distillation with the VAA module, \textsc{DeepFusion} effectively transfers the robust reasoning capabilities of local teacher models to the global MoE model.

\textbf{\textsc{DeepFusion} vs. Centralized MoE Training:} We evaluate the federated MoE training performance of \textsc{DeepFusion} by comparing it with a state-of-the-art centralized MoE training method named DeepSpeed. Our selected baseline  DeepSpeed is a well-established centralized MoE training method in academic and industrial settings, and provides a theoretical performance upper bound for our federated MoE training framework \textsc{DeepFusion}. Figure \ref{fig:deepspeed} illustrates the performance comparison between \textsc{DeepFusion} and DeepSpeed across two case studies with varying system scales. Results show that \textsc{DeepFusion} consistently achieves performance very close to that of centralized MoE training with DeepSpeed, demonstrating the effectiveness of our approach.

Note that for the medical multi-choice QA task, we use token accuracy as the evaluation metric, since the model's output tokens directly determine the correctness of multi-choice answers. For the financial open-ended QA task, we employ an LLM-as-a-judge evaluation framework \cite{Zheng2023Judging}, a widely-adopted method for assessing generative AI models on open-ended QA. Specifically, we leverage the Google Gemini 2.5 API to automatically score each response generated by the MoE model. The LLM judge assigns a score from 0$\%$ to 100$\%$, reflecting how well the response addresses the posed question.

\section{Related Work}\label{sec:related_work}
\subsection{Federated MoE Training}
Federated learning addresses the extensive training data needs of MoE models by enabling access to private data distributed across countless edge devices. Recent research \cite{Guo2021PFL,Dun2023Fedjets,Zhan2024FedMoE,Feng2025PMMOE} has made some attempts to federated MoE training, primarily by leveraging \textit{expert parallelism} across edge devices. Specifically, each edge device is assigned to train a pruned MoE model as \textit{a local expert model} in parallel using its private data. After each training round, edge devices transmit their trained local expert models to a central server, which aggregates them to construct the global MoE model.

However, existing studies often require edge devices to train assigned local expert models which are essentially pruned MoE models. Their backbone MoE architecture remains overloaded for resource-limited edge devices, preventing broad device participation in federated MoE traning. Additionally, these studies focus on small-scale, custom-built MoE architectures for computer vision tasks. In contrast, our approach leverages ubiquitous edge devices from tiny AIoT nodes to autonomous vehicles, regardless of hardware limitations, to contribute local expertise to the global MoE model. Furthermore, our proposed solution targets industry-level MoE-based LLMs, underscoring its practical applicability.

\subsection{Federated Knowledge Distillation}
Federated knowledge distillation \cite{Eldar2022Federated} has gained attention for supporting heterogeneous edge devices in federated learning. Each edge device independently configures and trains its local model based on its hardware capabilities. Then, these trained local models (teachers) share knowledge via soft-label predictions (logits) \cite{Xia2024AeroReC, Fan2025FedMKT} or intermediate features \cite{Yang2023Fed, Shao2025FedUFD}. The logits-based method trains the global model (student) to align its final soft-label predictions with those of the local models. The feature-based method leverages intermediate features from teachers as hints to guide the student in learning the teachers' feature representations, enabling more efficient knowledge transfer. These learned feature representations capture the teachers' internal data analysis and reasoning capabilities.

However, most current methods simply assume that teacher and student models share homogeneous architectures (e.g., Qwen1.5-0.5B vs. Qwen1.5-1.8B \cite{Bai2023Qwen}). Knowledge transfer is conducted through direct teacher-student logits or feature alignment in a shared latent space. To our knowledge, FedKMT \cite{Fan2025FedMKT} is the first to explore federated knowledge transfer between large and small language models, closely aligning with our work. The global LLM performance is improved by transferring domain insights from local small models. However, this work does not target MoE-based LLMs, nor does it identify the view-mismatch issue in federated knowledge distillation between heterogeneous model architectures (e.g., large and small language models). In our work, we address this issue by proposing a VAA module to mitigate latent space misalignment during cross-architecture federated knowledge distillation.

\section{Conclusion}
In this paper, we propose \textsc{DeepFusion}, a novel federated MoE training framework that enables heterogeneous resource-constrained edge devices to join MoE training via federated knowledge distillation. By allowing each device to independently configure and train an on-device LLM, \textsc{DeepFusion} aggregates diverse local domain knowledge from a spectrum of on-device LLMs, significantly enriching the global MoE model's capabilities. We address the critical challenge of cross-architecture knowledge distillation by designing the View-Aligned Attention (VAA) module, which aligns the latent representations and predictive perspectives between heterogeneous model architectures, specifically on-device LLMs (teachers) and global MoE model (student). Experiments on real-world medical and financial datasets using industry-level MoE-based LLMs demonstrate that \textsc{DeepFusion} outperforms key baselines, particularly in open-ended QA tasks. Overall, \textsc{DeepFusion} enables scalable, privacy-preserving federated MoE training on heterogeneous, resource-constrained devices, addressing both the shortage of high-quality public data and the MoE model's extensive training data requirements.

\ifCLASSOPTIONcaptionsoff
  \newpage
\fi



%\bibliographystyle{IEEEtran}

\newcommand{\BIBdecl}{\setlength{\itemsep}{0.12 em}}
\bibliographystyle{IEEEtran}
\bibliography{./manuscript}

\end{document}